\documentclass[11pt]{article}

\usepackage[final]{acl}

\usepackage{times}
\usepackage{latexsym}
\usepackage{url}
\usepackage{booktabs}
\usepackage{longtable}
\usepackage{multirow}
\usepackage{array}
\newcolumntype{P}[1]{>{\raggedright\arraybackslash}p{#1}}
\usepackage{enumitem}

\usepackage[T1]{fontenc}

\usepackage[utf8]{inputenc}

\usepackage{microtype}

\usepackage{inconsolata}

\usepackage{graphicx}
\usepackage{framed}
\usepackage{fvextra}
\usepackage{float}
\usepackage{fvextra}
\usepackage{float}
\usepackage{amsmath,amssymb}
\usepackage{arydshln}
\usepackage{xcolor}

\renewcommand{\texttt}[1]{\emph{#1}}

\begin{document}
\title{Beyond Accuracy: Community Perspectives on Machine Translation}

\author{ 
    Yujun Wang$^{\clubsuit}$ \, 
    Ehud Reiter$^{\clubsuit}$ \,
    Shimei Pan$^{\diamondsuit}$ \,
    Steffen Eger$^{\spadesuit}$ \,
    Wei Zhao$^{\clubsuit}$ \,
    \\[0.4em]
    $^{\spadesuit}$University of Technology Nuremberg, Germany \\
    $^{\diamondsuit}$University of Maryland, Baltimore County, USA \\[0.4em]
    $^{\clubsuit}$ The Aberdeen NLP Research Group\\
    University of Aberdeen, UK\\[0.4em]
    Project website: \href{https://beyond-accuracy.nlp4sci.com/}{https://beyond-accuracy.nlp4sci.com/}\\ [0.4em]
    \href{mailto:y.wang2.25@abdn.ac.uk}{y.wang2.25@abdn.ac.uk} \ \ 
    \href{mailto:wei.zhao@abdn.ac.uk}{wei.zhao@abdn.ac.uk}
  }

\maketitle
\begin{abstract}
Despite remarkable progress in machine translation (MT), non-AI communities have raised growing concerns about MT systems, suggesting a noticeable gap between technical advancement and the needs of real-world users. For instance, while NLP researchers focus on benchmark performance, end users care about ethical concerns, trust, reliability, costs, and more. We argue that listening to various user communities is essential so that research efforts would be directed towards the problems that the communities care about. To this end, we present a large-scale analysis, for the first time, that investigates what four stakeholder communities (AI developers, professional translators, language learners, and language service providers) post about MT technology on social media. To do so, we construct a dataset of 79,286 posts and comments from Reddit, Facebook, Bluesky, and Mastodon from 2019 to 2025, and analyse where these communities disagree, and how and why. Overall, we find that communities often disagree, and even show strong conflicts due to polarised sentiments on topics such as translation quality, efficiency, and reliability. This is because these communities approach these topics differently: the AI community frames them as technical and computational problems, while non-AI (user) communities care more about quality nuances, time savings, user trust, and broader social issues.

\end{abstract}

\section{Introduction}
\begin{figure}[t]
  \centering
  \includegraphics[width=\linewidth]{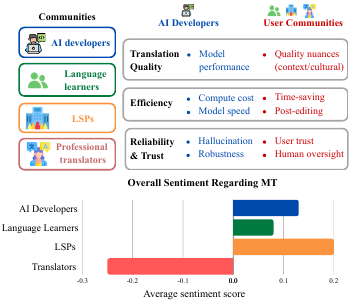}
 \caption{
 Different communities share common interests in topics such as translation quality, efficiency, and reliability, but frame them differently. For instance, AI developers relate translation quality to model performance, while translators relate it to quality nuances.
 }
  \label{fig:fig8}
\end{figure}
There have been decades of effort in developing machine translation (MT) systems, with NLP researchers  pushing the boundaries of MT to achieve near-human capabilities. However, MT is not only a topic of interest in the NLP community, which has become a core infrastructure that multiple communities rely on. For instance, professional translators use MT as a support tool to save time through post-editing machine translations
\cite{domingo2019incremental}.
Language departments use MT as a pedagogical tool to support language learning and translation training \citep{kirchhoff2024machine,kruk2025investigating}. Language Service Providers (LSPs) rely on MT to provide multilingual language services for clients \citep{presas2016machine}. Despite the diverse stakeholder groups of MT systems, their beliefs about what constitutes a ``good'' MT system seem not the same. For instance, AI researchers may associate a high-quality MT system with strong performance on benchmark datasets \citep{wu2025perhaps}; professional translators may consider a good MT system to be one that meets the standards of expert translators \citep{kruger2022some}; language learners may prioritise the usability of MT systems \citep{nino2020exploring}.

Such belief differences would lead to disagreements between communities regarding which MT system best suits their needs (or whether any system does).
The consequences of these disagreements  include low adoption rates of MT systems \citep{rivas2025translators}, distrust from users
\citep{sha2025understanding,man2026predictive}, and more.

Recently, there have been several critiques of MT from different communities. For instance, professional translators have raised concerns about job displacement, deskilling, and unfair compensation, high environmental costs \citep{rivas2024exploring}, while language learners
express their needs for the wider coverage of languages and dialects \citep{blaschke2024dialect}.
The AI community questions the ability of MT systems to handle the creativity and cultural nuances in literary translation \citep{van2026my, zhang2026beyond}, and many more.
Prior work has studied MT regarding technical challenges from the AI community \citep{castilho2025survey, karpinska2023large, zhang2025good}, and sociotechnical aspects of MT technology from non-AI communities \citep{bowker2019machine, rivas2024exploring, jimenez2025human}.
However, this line of work has several limitations:
(i) there are very few cross-community studies that discuss where communities disagree, how and why; (ii) no metric exists to measure the intensity of conflict within communities; (iii) it remains unclear what conflicts look like within and across communities;
(iv) there is a lack of community analysis across neural MT and LLM-based systems, and language pairs.
In this work, we present a cross-community study to address the aforementioned issues.
Our contributions are outlined below:

\begin{enumerate}[label=(\roman*)]
\setlength\itemsep{-0.4em}
    \item We construct a dataset of 79,286 MT-related posts and comments on 4 social media platforms (Reddit, Bluesky, Facebook, and Mastodon) from 2019 to 2025, covering 4 stakeholder communities: AI developers, language learners, language service providers, and professional translators.
    \item We use both LLMs and human annotators to analyse posts and comments across several dimensions: which stakeholder community a post or comment belongs to, whether the sentiment is positive or not, what topic and which MT system are discussed, among others. Additionally, we present a simple sentiment-based metric to measure the degree of community conflicts.
    \item We find that conflict within communities is not static, but evolves over time through expansion into new areas and (de-)escalation on existing ones. Additionally, we showcase how the topics of interest within the community of professional translators change over time, as well as their accompanying conflicts.
    \item We find that while different communities share common interests such as translation quality, efficiency, and reliability, they approach these topics differently: for instance, in terms of efficiency, AI developers care about compute cost and model speed, while translators focus on time savings from post-editing machine translations. We also break down our analysis into language pairs, and neural MT and LLM-based systems.
\end{enumerate}

\section{Related Work}

\paragraph{Community studies on machine translation.}
MT has been studied in both AI and non-AI communities, but from very different perspectives. From the non-AI side, translation scholars have investigated how MT is adopted, perceived, and integrated into professional translation workflows. For instance, prior work has examined topics such as MT literacy \cite{bowker2019machine}, ethical concerns \cite{rivas2024exploring}, translator education \cite{alfredo2024human}, and the changing role of human translators in AI-assisted translation workflows \cite{jimenez2025human}, focusing on the sociotechnical aspects of MT rather than the technology itself.
From the AI side, researchers focus on tackling technical challenges in MT, spanning topics such as low-resource translation, bias, evaluation metrics, long-context translation, and, more recently, literary translation, among many others \cite{castilho2025survey, shen2024survey}. For instance, \citet{karpinska2023large} find that more recent LLMs can effectively leverage document-level context, e.g., they translate better at the paragraph level than at the sentence level, but still commit critical omissions as well as grammatical errors. More recent work has highlighted issues in human evaluation, e.g., the gap between expert and non-expert human annotators, and has proposed new evaluation metrics that attempt to better approximate expert human judgments \cite{zhang2025good, zhang2025litransproqa, zhang2026beyond}. To our knowledge, no prior work has conducted a cross-community study on MT. This is the gap that we fill.

\paragraph{Social media studies on public opinions.} AI/NLP researchers have leveraged social media data to investigate public opinion across a wide range of domains, such as health and pandemics \cite{feldman2021analyzing,ding2017multi}, environmental issues \cite{che2025using}, AI perception \citep{leiter2024chatgpt}, and many more. For instance, prior work has analysed Reddit posts to identify toxic comments regarding online harassment \cite{almerekhi2022investigating}, and tracked emotional responses to climate change \cite{shaeri2025sentiment}. Some studies adopt a multi-platform approach, collecting data from Reddit, YouTube, and Twitter to incorporate diverse user demographics and communication styles \cite{amangeldi2024understanding}. Additionally, we note that social scientists have leveraged social media data for community studies, comparing the views of different stakeholder groups in domains such as finance \cite{gomez2021stakeholders} and healthcare \cite{holtorf2023using}. Our work builds on this line of research, but focuses on the MT context.

\section{Our Corpus}
Our dataset consists of posts and comments regarding MT, which we collected from four free-access social media platforms: Reddit, Bluesky, Facebook, and Mastodon\footnote{Since 2023, X/Twitter has ceased supporting free API access to its data.},  covering the time span from February 2019 to December 2025. We chose early 2019 as the start time because it is arguably the moment when the capabilities of MT systems are largely improved due to the rise of large language models.

\paragraph{Data collection.}
We collected posts and comments of relevance to MT using the same set of manually curated search queries
that combine (i) AI-related terms, e.g., LLMDevs, localLLM, (ii) translation-related terms, e.g., translation studies, translators and (iii) popular MT systems,
e.g., DeepL, ChatGPT. For Reddit, we targeted
relevant subreddits and applied search queries only to them;
for Bluesky and Mastodon, we applied search queries in the form of keywords and hashtags;
for Facebook, we relied on keyword search.
Full platform-specific query lists (subreddits, keywords, hashtags, etc) can be found in Appendix~\ref{sec:platform_keywords}. Our data statistics are provided in Table~\ref{tab:dataset_stats}.

\begin{table}[t]
\centering
\footnotesize
\begin{tabular}{llr}
\toprule
\textbf{Type} & \textbf{Category} & \textbf{Value} \\
\midrule
Overall & Time span & 2019--2025 \\
 & Total records & 105{,}310 \\
\midrule
Platform & Facebook & 78{,}401 \\
 & Reddit & 18{,}106 \\
 & Bluesky & 6{,}929 \\
 & Mastodon & 1{,}874 \\
\midrule
Community & AI developers & 31{,}456 \\
 & Language learners & 23{,}877 \\
 & Language service providers & 17{,}435 \\
 & Professional translators & 6{,}518 \\
 & \textbf{Total} & 79{,}286 \\
\bottomrule
\end{tabular}
\caption{Data statistics of our corpus.
}
\label{tab:dataset_stats}
\end{table}

\begin{figure}[t]
  \centering
  \includegraphics[width=\linewidth]{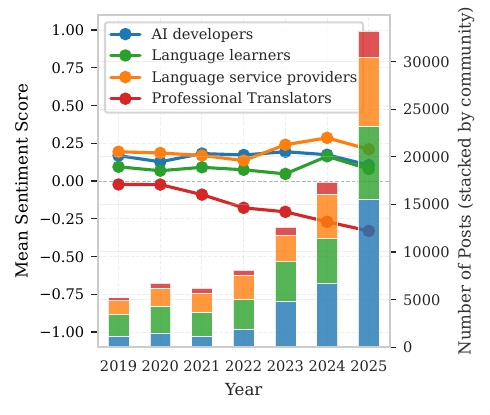}
  \caption{
  Data distribution (bar chart) and average (mean) sentiment score per community over time (line plot). Professional translators show a noticeable sentiment decline over time. Appendix~\ref{sec:appendix_platform_time} breaks down the data distribution into social media platforms.
  }
  \label{fig:fig1}
\end{figure}

\paragraph{Unit of analysis.}
Our analysis focuses on social media content: Facebook posts and comments from the Meta content library, Reddit posts and comments, Bluesky posts and comments, Mastodon posts retrieved via hashtag/keyword queries. Posts and comments are treated independently, meaning that we discard relationships between them to maximise our dataset volume, but we acknowledge that doing so causes comments to lose the context of their parent posts. In the following, we use an LLM to annotate every post/comment in several dimensions: community, sentiment, subject, and aspect, among others. Human evaluation of the LLM annotations is provided thereafter.

\paragraph{Community classification.} A ``community'' is defined as a stakeholder group that posts opinions about MT on social media platforms. Our work focuses on four communities: AI developers, language learners, LSPs, and professional translators. Each post/comment is assigned one community label. If there is no clear indication of which community a post/comment belongs to, or if it belongs to a novel stakeholder group, it is assigned to ``others''.
We use GPT-4o-mini to perform
single-label community classification. For each platform,
we extract the primary text field, strip whitespace, drop empty entries, and prompt the LLM to assign a community label to each text segment.
We removed spam/ads and obviously off-topic posts/comments (see Appendix~\ref{sec:appendix_community_label} including our prompt).

\paragraph{Sentiment classification.}
We use the sentiment classifier from \citet{barbieri2022xlm} to predict the sentiment label of each text segment. The classifier, built upon XLM-RoBERTa \citep{conneau2020unsupervised}, has been trained on approximately 200 million tweets and fine-tuned on Twitter sentiment datasets. We choose this model because our data also consists of informal, user-generated social-media posts, and more importantly, this classifier has been widely applied to Facebook \cite{zhang2025public} and Reddit content \cite{piorino2025sentiment}.
Here, each comment/post is assigned a sentiment label (negative, neutral, or positive). We acknowledge that three sentiment labels cannot fully represent
the sentiment of a post/comment, as it may contain mixed sentiments or sarcasm (i.e., the sentiment may be negative even if it sounds positive).

\paragraph{Other dimensions of annotation.}
We further use GPT-4o-mini to automatically annotate posts and comments in other dimensions: (i) the subject being discussed (e.g., a specific MT system), (ii) an aspect of the subject (e.g., features of a specific MT system, such as translation quality), (iii) a key statement regarding the subject (in the VERB + OBJECT format, e.g., ``improve translation quality''), and (iv) an explanation of the statement (e.g., explaining why translation quality is improved). Then, we group posts and comments if their aspects are similar. To do so, we manually define 15 broader aspect categories and map the GPT-annotated aspects to these categories; for instance, the ``Efficiency'' category includes aspects such as ``speed'', ``latency'', ``efficiency'', ``throughput'', and ``response time''. Our prompt is provided in Appendix~\ref{sec:appendix_claim_aspect}.

\paragraph{Human evaluation.}
\begin{table}[t]
\centering
\small
\begin{tabular}{lcc}
\toprule
\textbf{Dimension} & \textbf{Inter-Annotator} & \textbf{Accuracy} \\
    & \textbf{Agreement}  & \\
\midrule
Community label & 0.725 & 0.82 \\
Sentiment label & 0.969 & 0.97 \\
Aspect & 0.953 & 0.95 \\
Verb--object & 0.988 & 0.99 \\
\bottomrule
\end{tabular}
\caption{
Evaluation of LLM results.
}
\label{tab:human_eval}
\end{table}
We manually evaluate the results regarding community and sentiment classifications, as well as the automated annotations regarding subjects, aspects, and verb-object statements. To do so, we use a stratified set of 500 posts and comments to evaluate community classification (100 per community across 5 communities including the ``Others'' community) and 400 posts/comments (excluding ``Others'') to evaluate sentiment classification and annotations of subject, aspect, and verb-object statements. We involved two annotators (PhD students---one is from our team while another is not) to independently
annotate our subsets, then resolve any disagreements if any. Table ~\ref{tab:human_eval} shows the accuracy of LLM classifiers/annotators across all dimensions, as well as inter-annotator agreement. We find that accuracy and human agreement exceed 90\% across all dimensions except for community labelling, for which we explain in our limitations section. Detailed breakdowns by community and social media platform are provided in Appendix \ref{sec:appendix_eval_detail}.

\section{Results}
\label{sec:analysis}

We analyse conflict in two ways: (i) within-community conflict, where the community disagrees internally, and (ii) cross-community conflict, where various communities disagree.

\begin{table}[t]
\centering
\footnotesize
\setlength{\tabcolsep}{3pt}
\renewcommand{\arraystretch}{1.12}
\begin{tabular}{p{0.18\columnwidth}p{0.74\columnwidth}}
\toprule
\textbf{Sentiment} & \textbf{Posts and Comments} \\
\midrule

Positive
& \emph{``language translation is something these models are actually well-suited for''} \\ \cdashline{2-2}
& \emph{``Speech translation systems can be more robust to background noise \ldots and achieved results that outperformed other leading translation models''} \\

\midrule

Negative
& \emph{``ChatGPT output should not be used for \ldots translation without verification''} \\\cdashline{2-2}
& \emph{``if the context is too large, they can leave out some sentences \ldots these LLMs can't translate reliably''} \\

\bottomrule
\end{tabular}
\caption{Post/comments snippets from AI developers.
}
\label{tab:ai_dev_reliability_mixed}
\end{table}
\subsection{Within-Community Analyses}

\begin{figure*}[t]
  \centering
  \includegraphics[width=\textwidth]{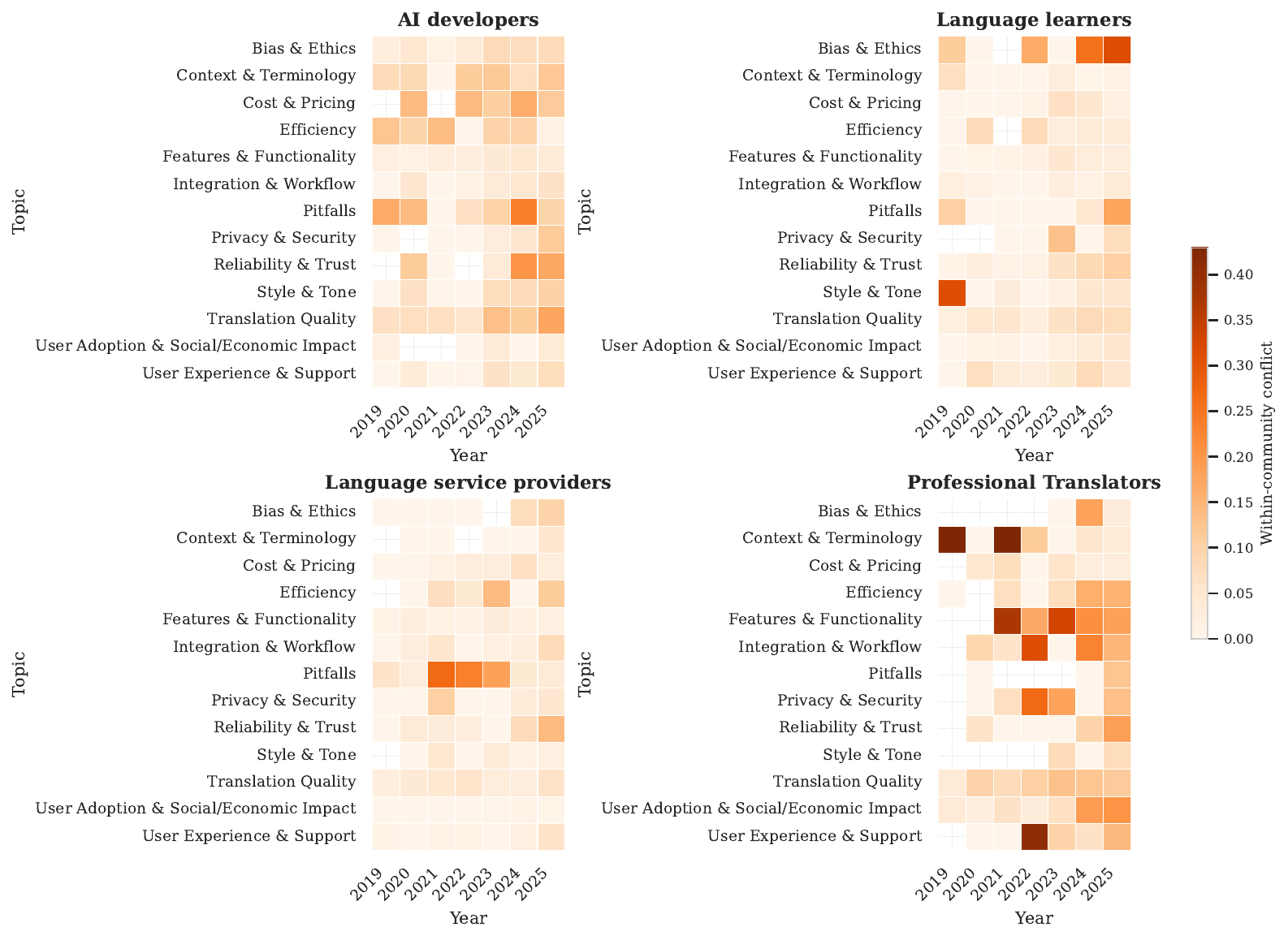}
  \caption{Within-community conflict intensity by topic and year. Each panel corresponds to one community. Rows show topics (top 10 by volume),
  The darker the colour, the more intense a community in that topic/area.
  We do not colour cells where fewer than five posts/comments are available.}
  \label{fig:fig2}
\end{figure*}
Table \ref{tab:ai_dev_reliability_mixed} showcases what AI developers disagree about regarding MT systems. To examine the degree to which a stakeholder group is internally divided, we design a metric that measures the conflict intensity per community. The metric multiplies two components: the degree to which sentiments are spread out (standard deviation) and the degree to which positive and negative sentiments are mixed. We apply this metric to compute an intensity score for each area per year, where a higher score indicates stronger conflict intensity. Details of our metric is provided in Appendix~\ref{sec:appendix_conflict_metrics}.

\paragraph{Conflict expansion and escalation.} Figure~\ref{fig:fig2} shows conflict intensity scores across communities.
We observe that conflict within communities grows over time in two ways: spreading to new conflicting areas (expansion) or intensifying on existing  areas (escalation). For instance, conflict among professional translators expands to new areas, such as Style \& Tone (post-2023) and Features \& Functionality (post-2021), while conflicts regarding Efficiency, User Adoption \& Social Impact are generally escalated over years (with the colours getting darker over time); we also note that some conflicts de-escalate over time, e.g., Context \& Terminology. AI developers show expansion into areas that were previously dormant but revived later on, such as Reliability \& Trust (post-2023) and Social/Economic Impact (post-2022); they show conflict escalation in Translation Quality, and Privacy \& Security.
For language learners and LSPs, we find some areas are highly intensive, such as Bias \& Ethics, and Pitfalls. We also note that professional translators seem to show the strongest conflict, meaning that they are the most divided community. This is quite interesting,
given growing criticisms of MT from translators. Table \ref{tab:pro_translator_mixed} showcases data with mixed sentiments from translators, suggesting that many of them are positive about MT (e.g., productivity gains and translation quality in some domains).

\begin{table}[t]
\centering
\footnotesize
\setlength{\tabcolsep}{3pt}
\renewcommand{\arraystretch}{1.12}
\begin{tabular}{p{0.18\columnwidth}p{0.74\columnwidth}}
\toprule
\textbf{Sentiment} & \textbf{Posts and Comments} \\
\midrule
Positive
& \emph{``Neither is good enough without careful checking and rewriting, but they do enable me to work much faster''}\\ \cdashline{2-2}
& \emph{``I translated the lyrics to a Spanish rap song \ldots and GPT gave me a fine translation. It only needed a couple edits''} \\ \cdashline{2-2}
\midrule
Negative
& \emph{``AI models like DeepL are still almost completely useless. With complex literary language, they just fail''} \\ \cdashline{2-2}
& \emph{``Do not use Google to translate Arabic to English when complex meanings are involved''} \\
\bottomrule
\end{tabular}
\caption{Post/comment snippets
from translators.}
\label{tab:pro_translator_mixed}
\end{table}

\paragraph{Topic shift over time.} Even if conflict recurs in the same area over time, that does not mean that the issues encountered within the community are the same. For instance, we manually analysed posts and comments regarding Features \& Functionality from professional translators. Table \ref{tab:features_functionality_timeline} summarises topic shifts and tensions among translators. We find that translators often discuss CAT infrastructure (2021), then switch to AI integration (2023), and to automation of translation workflow (2025). Overall, we observe that the conflicting issues differ by year, but generally have the same divide: the desire for greater efficiency and automation on the one hand, and concerns about quality, reliability, and human oversight on the other.

\begin{table}[t]
\centering
\small
\setlength{\tabcolsep}{3pt}
\renewcommand{\arraystretch}{1.12}
\begin{tabular}{@{}p{0.11\columnwidth}p{0.39\columnwidth}p{0.42\columnwidth}@{}}
\toprule
\textbf{Year} & \textbf{Main Focus} & \textbf{Key Tension} \\
\midrule
2021
& CAT core infrastructure, including TM, segmentation, and termbases.
& Dependence on stable core functions vs.\ concerns about TM/TB failures. \\

2022
& Tool adoption and usability in increasingly complex CAT suites.
& Feature-rich functionality vs.\ steep learning curves and clunky interfaces. \\

2023
& AI integration in CAT workflows, including MT plugins and AI-assisted translation.
& Automation embedded in CAT tools vs.\ translators' ability to control and refine MT output. \\

2024
& Knowledge management, including contextual phrase databases and terminology handling.
& Demand for richer context support vs.\ added system complexity and maintenance burden. \\

2025
& Workflow automation and QA, including automated checks, extraction tools, and API pipelines.
& End-to-end efficiency and automation vs.\ accuracy, reliability, and professional oversight. \\
\bottomrule
\end{tabular}
\caption{
Topic shifts and tensions among translators in the area of Features \& Functionality.
}
\label{tab:features_functionality_timeline}
\end{table}

\subsection{Cross-Community Analyses}
\label{sec:cross-community}

\begin{figure*}
  \centering
  \includegraphics[width=\textwidth]{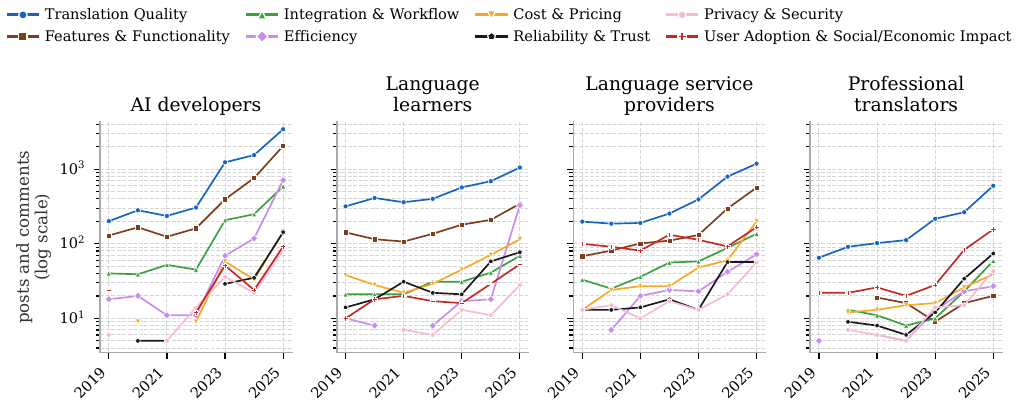}
  \caption{Data volume by topic and community over time (2019–2025).
  }
  \label{fig:fig3}
\end{figure*}

\paragraph{Which topics of interest differ across communities?}
Figure~\ref{fig:fig3} shows which topics grow over time. We find that some topics are of common interest to all communities, but their momentum differs. For instance, Translation Quality grows slowly for language learners compared to the other three communities, which might be because language learners, who are not experts in translation studies, do not have the expertise to evaluate translation quality. We also find that Privacy \& Security seems a long-standing interest for LSPs, with posts and comments throughout all years, while the other three communities discuss this topic only periodically. This is understandable, because LSPs, which serve a large user base, are quite sensitive to privacy concerns.
We also find that User Adoption \& Social/Economic Impact is most discussed by professionals (esp. LSPs and Translators), perhaps because they are directly affected by the social and economic consequences of MT tools.

\begin{figure}[!t]
  \centering
  \includegraphics[width=\linewidth]{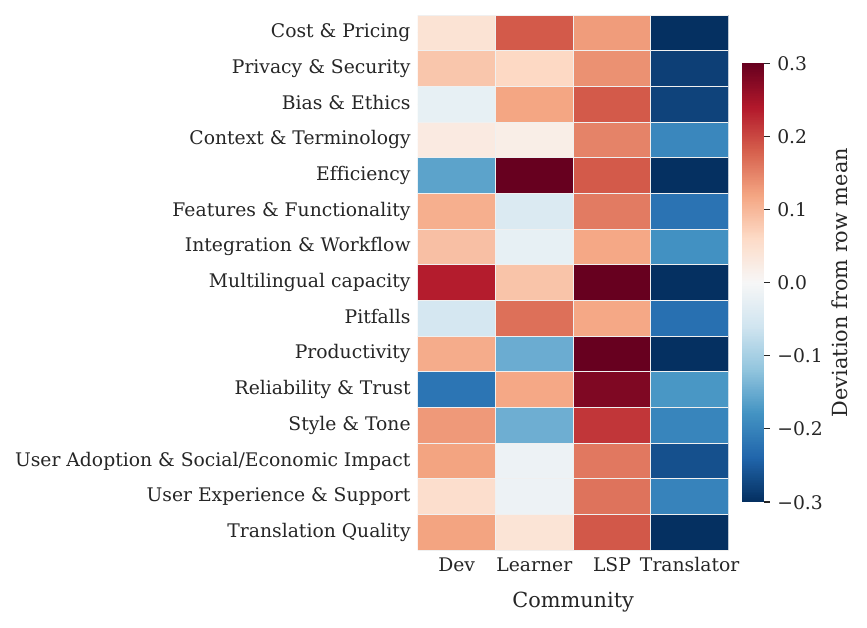}
  \caption{``Normalised'' sentiment scores across communities. Cell colour indicates whether a community is more positive/negative than average on one topic.}
  \label{fig:fig5}
\end{figure}

\paragraph{How does sentiment differ across communities?}
To make sentiment scores comparable across communities, for each topic we compute the average sentiment score per community, then remove the overall average on that topic. We call this a ``normalised'' sentiment score, showing whether a community is more positive/negative than average on the topic.
Figure \ref{fig:fig5} shows normalised sentiment scores. Overall, we find that LSPs are the most positive across all areas compared to other communities, while professional translators are the most critical. This suggests that translators are generally dissatisfied with current MT systems, e.g., Bias \& Ethics.
LSPs being the most positive may be because they do not critically evaluate MT systems but simply adopt them into their workflows. Language learners show neutral sentiment regarding productivity, as they likely do not use MT systems to support their jobs. AI developers are generally positive in most areas, e.g., Style \& Tone---the area largely criticised by translators. This may suggest that AI developers who lack expertise in translation studies are less critical than translators.

\begin{table*}[t]
\centering
\footnotesize
\setlength{\tabcolsep}{3.5pt}
\begin{tabular}{@{}p{0.17\textwidth}p{0.25\textwidth}p{0.34\textwidth}p{0.18\textwidth}@{}}
\toprule
\textbf{Topic} &
\textbf{AI Community (subtopics)} &
\textbf{Non-AI Community (subtopics)} &
\textbf{Framing Gap} \\
\midrule

\textbf{Translation Quality} &
Accuracy and adequacy; context, nuance, and register; human review and quality assurance &
Human review and quality assurance; accuracy and adequacy; context, nuance, and register &
Measurable quality vs. accountable quality \\

\textbf{Efficiency} &
Cost, tokens, and compute resources; workflow integration; model optimization and inference efficiency &
Workflow integration; MTPE / post-editing workflow; quality control and review burden &
System efficiency vs. human workload \\

\textbf{Reliability \& Trust} &
Accuracy, hallucination, and factual trust; trust, adoption confidence, and reputation; stability, outages, and bugs &
Trust, adoption confidence, and reputation; accuracy, hallucination, and factual trust; verification, review, and human oversight &
Technical reliability vs. practical trust \\

\bottomrule
\end{tabular}
\caption{Gaps between the AI community (AI developers) and non-AI communities (language learners, LSPs, and translators).
For topic, we list subtopics from both AI and non-AI communities, based on our corpus (2023--2025).
}
\label{tab:subtopic_framing_gap}
\end{table*}

\paragraph{Why does sentiment differ across communities?}
We manually analysed posts and comments to investigate why sentiment on the same topics differs across communities. Table \ref{tab:subtopic_framing_gap} provides subtopics that we identified from the AI community (AI developers), as well as the non-AI communities (language learners, LSPs, and professional translators). Overall, we find that the two communities seem to exhibit different values, needs, and interests. For instance, both communities care about translation quality, efficiency, and reliability, but define these concepts differently: AI developers associate (i) translation quality with model performance, (ii) efficiency with compute cost, inference speed, and model optimisation, and (iii) reliability with hallucination, factuality, robustness, and system outages. In contrast, the non-AI community refers translation quality to nuanced quality dimensions regarding context, terminology, register, and thinks about efficiency in terms of time reduction in post-editing machine translation compared to human translation, and associates reliability with confidence in adopting MT tools, human verification/oversight, and the accountability of MT systems. For these reasons, it is not surprising that the AI and non-AI communities sometimes hold different sentiments, and even conflicting viewpoints regarding MT systems. More details are provided in Table \ref{tab:subtopic_framing_gap_full}.

\begin{table}[t]
\centering
\footnotesize
\setlength{\tabcolsep}{2.5pt}
\begin{tabular}{llccc}
\toprule
\textbf{Community} & \textbf{System} & \textbf{Pos.} & \textbf{Neg.} & \textbf{Diff.} \\
\midrule

AI dev.
& NMT ($n$=4,949) & 28.25 & 7.13 & \textbf{21.12} \\
& LLM ($n$=4,907) & 27.17 & 19.34 & 7.83 \\\midrule

Learners
& NMT ($n$=4,368) & 16.16 & 12.25 & 3.92 \\
& LLM ($n$=918) & \textbf{44.01} & 14.27 & \textbf{29.74} \\\midrule

LSPs
& NMT ($n$=2,529) & 23.41 & 7.63 & 15.78 \\
& LLM ($n$=389) & 26.48 & 12.34 & 14.14 \\\midrule

Translators
& NMT ($n$=414) & 10.39 & 31.16 & \textbf{$-20.77$} \\
& LLM ($n$=211) & 15.64 & \textbf{41.23} & $-25.59$ \\
\bottomrule
\end{tabular}
\caption{Sentiment regarding NMT and LLMs. Pos. and Neg. are percentages; Diff. is Pos. minus Neg.}
\label{tab:nmt-llm-sentiment}
\end{table}

\begin{figure*}[t]
  \centering
  \includegraphics[width=\textwidth]{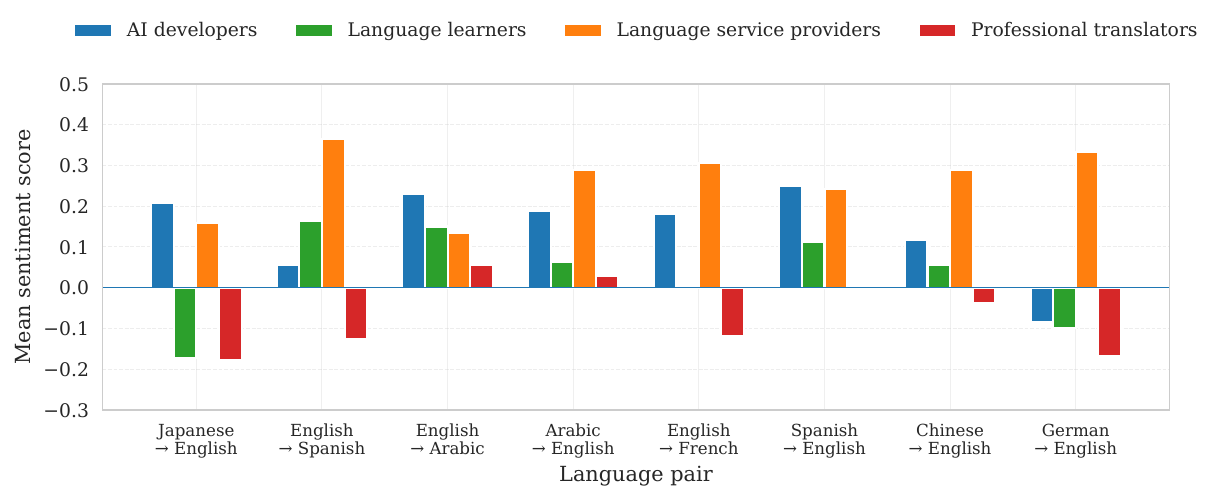}
  \caption{Average (mean) sentiment score per community across 8 language pairs.
  }
  \label{fig:fig7}
\end{figure*}

\paragraph{NMT and LLM-based systems.}
We break down our sentiment analysis into two system types: (a) NMT systems: Google Translate, DeepL, LibreTranslate, and ImTranslator, and (b) recent LLM-based systems: GPT, Claude, Meta AI, Google LLM, Microsoft Copilot, Grok, and DeepSeek. Table \ref{tab:nmt-llm-sentiment} shows how sentiment differs between these two types. We find that professional translators show the largest negative sentiment, with stronger negativity for LLM-based systems than NMT systems, meaning that translators' concerns may have been intensified by recent LLMs (a greater threat to their jobs).
AI developers and LSPs are  more negative about LLMs than NMT systems; this might be because, unlike NMT, LLMs are general-purpose models not  tailored for translation tasks. Language learners show the strongest positive sentiment for LLMs, perhaps because they benefit from the greater usability of these models, allowing for tasks like conversation-based machine translation.

\paragraph{Language pairs.}
Figure \ref{fig:fig7} shows the average sentiment score per community across 8 language pairs. We identify language pairs in posts and comments by searching for translation-direction patterns such as "X to Y", "X--Y", "from X to Y", and "translate X into Y". This subset covers several translation topics such as regional dialects, cross-script translation (Latin to non-Latin script), cultural and technical contexts, among others. We find that communities are generally positive about Arabic$\leftrightarrow$English, Spanish$\rightarrow$English, and Hindi$\rightarrow$English translation. However, sentiments differ across communities in Japanese$\rightarrow$English, English$\rightarrow$Spanish, and English$\rightarrow$French. To understand the mixed sentiments, we provide case studies (see Table \ref{tab:language-pair-case-studies}): For Japanese$\rightarrow$English, AI developers are positive, acknowledging accuracy improvements, while professional translators are negative, criticising DeepL that fails to handle the complex literary language. For German$\rightarrow$English, LSPs are positive, e.g., impressed by DeepL's quality, while other communities are negative (grammar and capitalisation errors in Google Translate, and poor performance in casual and creative contexts).

\section{Conclusion}
In this paper, we analysed how stakeholder communities discuss machine translation on social media platforms. Among the four communities, we showed that professional translators are the most negative regarding quality, cost, trust, and labour impacts. We further examined which topics of interest differ across communities, how and why they differ, how conflicts expand and escalate over time, and how user sentiments differ across neural MT and LLM-based systems and language pairs. Overall, we found that these differences stem from community-specific goals that could grow into intense  conflicts between communities over time.

Given these conflicts, MT tools cannot truly benefit all stakeholders unless their design meets the needs of all impacted communities. We therefore recommend that future evaluation of MT systems goes beyond quality metrics (e.g., BLEU), but also incorporates metrics that reflect trust, efficiency, labour impact, and human-AI translation workflow. Additionally, non-AI communities (professional translators, language service providers, and language learners) shall participate in the design and evaluation of MT tools so that conflicts can be de-escalated for collective benefits of all stakeholders.
\section*{Limitations}\label{sec:limitations}

Our data collection strategy is constrained by platform-specific query mechanisms, which limits the coverage of social medial data.
For instance, discussions in private groups, closed professional forums, and platforms with restricted
access models are
excluded by us. We also acknowledge the uneven distribution across platforms (Facebook: 78.3\%, Reddit: 16.8\%, Bluesky: 3.8\%, Mastodon: 1.1\%); due to data scarcity, platform-stratified analysis could be challenging; therefore, we do not investigate how public views regarding MT differ across social media platforms. Rather we collect data from as many free-access platforms as possible to reach a large user base.
Further, social media data is mostly in English, which is the focus of our work, while we acknowledge that low-resource minority groups are under-represented in our data as their posts are far fewer.
Additionally, our keyword-based approach for data collection is not optimal, as it fails to collect data that do not exactly match the keywords but are semantically relevant.
Another issue is that community classification errors may propagate into downstream sentiment and other analyses. We acknowledge this, but the error rate is very small on Reddit (ACC: 92.16\%) and Bluesky (ACC: 91.30\%). However, accuracy is much lower on Facebook (ACC: 72.83\%) because it contains a substantial amount of off-topic content, e.g., ads that look like normal data and bypass our filtering method (see Table \ref{tab:platform-community-accuracy}).

\section*{Ethical Considerations}
We collected only publicly accessible content through official, platform-provided access channels and in accordance with each platform's terms and data policies. Specifically, collection followed Reddit developer terms and DPA requirements\footnote{\url{https://redditinc.com/policies/data-api-terms}}; Bluesky public access policies; Meta Content Library API access terms (version 6.0, dated January 2026; data collection for this study was completed by the date given in \S3.1)\footnote{\url{https://doi.org/10.48680/meta.metacontentlibraryapi.6.0}}, with access reviewed via CASD; and Mastodon's signed HTTP request requirements for public content retrieval. We did not access private groups or bypass access controls. To reduce privacy risks, we do not release raw post texts or user identifiers; we report aggregated statistics and, when illustrative examples are needed, we do not quote identifiable strings to prevent de-anonymisation.
We retain identifiers only for internal deduplication.

\section*{Acknowledgments}
We thank the anonymous reviewers for their constructive feedback, which substantially improved this work. The first author is financially supported by the China Scholarship Council (CSC). We are grateful to Chen Zhu for her help with annotation. The NLLG group (UTN) 
gratefully acknowledges support
from the German Research Foundation (DFG) via
the Heisenberg Grant EG 375/5-1.

\bibliography{custom}

\clearpage
\twocolumn
\appendix

\section{Additional Notes Supporting Main-Body Claims}
\label{sec:appendix_main_body_support}

This appendix section provides concise supporting detail for the main-body interpretation in \S\ref{sec:analysis}. First, the communication-gap claim is supported by repeated verb--object asymmetries across multiple aspects (not only \emph{Translation Quality}), where developer discourse is predominantly optimization-oriented while translator and learner discourse is more risk- and consequence-oriented. Second, the within-community dynamics claim is supported by recurring concentration of high-conflict cells in a limited set of aspects (quality, efficiency/cost, reliability/trust), rather than broad uniform conflict across all topics.

These supplementary observations are intended to strengthen traceability between the narrative argument in the main body and the detailed diagnostics reported in later appendix sections (especially \S\ref{sec:appendix_conflict_metrics} and \S\ref{sec:appendix_within_community}).

\section{Data Collection Queries by Platform}
\label{sec:appendix}
\subsection{Platform-Specific Keywords for Data Collection}
\label{sec:platform_keywords}

Because different platforms support different query mechanisms and discourse
styles, we used platform-specific keyword sets for data collection. The overall
selection strategy was consistent across platforms, while keywords were adapted
to platform affordances and usage conventions.

\paragraph{Reddit.}
We collected posts using the following subreddit--keyword query pairs:
\begin{itemize}
  \item \texttt{TranslationStudies}: \texttt{AI}, \texttt{artificial intelligence}, \texttt{machine}, \texttt{machine learning}, \texttt{neural networks}, \texttt{DeepL}, \texttt{Google Translate}
  \item \texttt{translator}: \texttt{AI}, \texttt{artificial intelligence}, \texttt{machine}, \texttt{machine learning}, \texttt{DeepL}, \texttt{Google Translate}
  \item \texttt{ChatGPT}: \texttt{translation}, \texttt{machine translation}
  \item \texttt{OpenAI}: \texttt{translation}, \texttt{machine translation}
  \item \texttt{machinelearning}: \texttt{translation}, \texttt{machine translation}
  \item \texttt{LocalLLaMA}: \texttt{translation}, \texttt{machine translation}
  \item \texttt{LLMDevs}: \texttt{translation}, \texttt{machine translation}
  \item \texttt{LocalLLM}: \texttt{machine translation}
  \item \texttt{grok}: \texttt{translation}
  \item \texttt{ClaudeAI}: \texttt{translation}, \texttt{machine translation}
  \item \texttt{Gemini}: \texttt{translation}
  \item \texttt{GeminiAI}: \texttt{translation}
  \item \texttt{MistralAI}: \texttt{translation}
  \item \texttt{machinetranslation}: \texttt{N/A}
\end{itemize}

\paragraph{Bluesky.}
Due to the hashtag-driven and short-form nature of the platform, we used the following keywords and hashtags.

\noindent \textbf{Keywords:}
\texttt{deepl}, \texttt{google translate}, \texttt{machine translation}, \texttt{ai translation},
\texttt{artificial intelligence translation}, \texttt{machine learning translation},
\texttt{neural networks translation}, \texttt{llm translation}, \texttt{chatgpt translation},
\texttt{openai translation}, \texttt{gemini ai translation}, \texttt{deepseek translation},
\texttt{claude ai translation}, \texttt{ai literary translation}.

\noindent \textbf{Hashtags:}
\texttt{\#machinetranslation}, \texttt{\#googletranslate}.

\paragraph{Mastodon.}
Mastodon is not a single website but a federated network consisting of many independent servers. To collect data, we selected several popular servers, including \texttt{mastodon.social}, \texttt{mas.to}, \texttt{social.vivaldi.net}, and \texttt{mstdn.party} for the general domain, as well as \texttt{techhub.social} and \texttt{defcon.social} for the technology domain.
We used the following keywords:
\texttt{deepl}, \texttt{googletranslate}, \texttt{machinetranslation}, \texttt{aitranslation},
\texttt{LibreTranslate}, \texttt{Lingva}.

\paragraph{Facebook.}
Data collection relied on keyword searches within public posts and groups:
\texttt{machine translation}, \texttt{AI translation}, \texttt{LLM translation},
\texttt{neural machine translation}, \texttt{context loss translation},
\texttt{chatgpt translation}, \texttt{claude translation}, \texttt{deepl}, \texttt{google translate},
\texttt{mistral translation}, \texttt{llama translation}, \texttt{gemini translation},
\texttt{technical translation}, \texttt{translation prompt}.
\section{Prompt Templates for Annotation}

\subsection{Community Labeling Prompt}
\label{sec:appendix_community_label}
The following prompt was used to classify each entry into exactly one community category.

\begin{framed}
\begin{Verbatim}[breaklines,breakanywhere,formatcom=\rmfamily]
You are an expert text classifier.
Your task is to classify the given text into exactly ONE category,
based on the PRIMARY perspective or intent expressed in the text.
Categories:
1. AI developers
The text discusses AI systems, models, or tools specifically for
translation, localization, or language processing (e.g. NLP, MT),
from a development, technical, or design perspective.
Includes: building, evaluating, integrating, or improving such systems.
2. Language learners
The text focuses on learning or studying a language,
language education, practice, or learner experiences.
Includes questions, tips, struggles, or resources for learning languages.
3. Language service providers
The text is written from the perspective of offering language related services
such as translation, localization, interpretation, or language consulting,
often in a business or client-facing context.
4. Professional translators
The text reflects the perspective of someone doing translation
as a profession, including workflows, tools (e.g. CAT tools),
quality issues, pricing, career concerns, or industry experiences.
5. Unclear
Use this ONLY if:
- The text has no clear connection to language, translation, localization, or
language-related AI
- OR the content is too vague or contextless to infer any perspective
Rules:
- Always choose the BEST-FIT category, even if the text
  could partially fit more than one.
- Do NOT use "Unclear" for multi-topic or ambiguous posts
  if a dominant perspective can be inferred.
- Base your decision only on the provided text.
- Return ONLY the label name as plain text (no explanation).
\end{Verbatim}
\end{framed}

\subsection{Topical Relevance and Spam Filtering Prompt}
\label{sec:appendix_relevance_prompt}
After preprocessing (deduplication and removal of empty/removed or obviously automated content), the full dataset comprises 202{,}051 posts and comments. We then use an instruction-following LLM to flag spam, advertising, and obviously off-topic material (including \emph{Unclear} items whose text is unrelated to translation), yielding an analysis set of 105{,}310 posts and comments. To assess the reliability of this topical filter, we manually reviewed 100 LLM-flagged \emph{Unclear} items and agreed with exclusion decisions in 97\% of cases, corresponding to an exclusion accuracy of 0.97.

\begin{framed}
\begin{Verbatim}[breaklines,breakanywhere,formatcom=\rmfamily]
You are an expert content filter.
Your task is to decide whether the following social media item
is clearly about machine translation (MT), AI translation, or
translation workflows, or whether it is spam/advertising or
off-topic with respect to translation.
Label the text as exactly ONE of the following categories:
1. on_topic
   The main content is explicitly about translation, machine
   translation, AI translation tools or models (e.g. DeepL,
   Google Translate, ChatGPT translation), translation quality,
   translation workflows, translators' work, or language
   learning that directly involves translation tools.
2. off_topic
   The text is not about translation at all, or only mentions
   translation in a trivial way without discussing translation
   tools, quality, workflows, or translation-related concerns.
3. spam_or_ad
   The text is primarily spam, generic promotion, job ads,
   marketing, link farming, or unrelated news, even if it
   happens to contain translation-related keywords.
Rules:
- Focus on whether the main point of the text concerns
  translation or translation tools.
- If in doubt, prefer "off_topic" rather than "on_topic".
- Return ONLY the label name: on_topic, off_topic, or spam_or_ad.
\end{Verbatim}
\end{framed}

\subsection{Claim and Aspect Extraction Prompt}
\label{sec:appendix_claim_aspect}
The following prompt was used to extract structured semantic information from each entry.

\begin{framed}
\begin{Verbatim}[breaklines,breakanywhere,formatcom=\rmfamily]
You are a specialized Narrative Semantic Analyst.
Your task is to extract structured information from the text below while
strictly following all constraints.
Text:
{text}
Extraction Fields:
1. subject: The main entity being discussed (e.g., a tool, system, organization,
   or policy).
2. aspect: The specific feature or dimension of the subject that is explicitly
   discussed, praised, or criticized.
3. verb_object: A concise predicate describing the expressed claim, using a
   VERB + OBJECT format (e.g., "improve efficiency", "reduce quality").
4. explanation: A single-sentence explanation summarizing the rationale behind
   the expressed sentiment or opinion.
If any field cannot be inferred from the text, output the value "N/A".
Return ONLY a valid raw JSON object.
Do NOT include markdown formatting, code fences, or additional text.
Expected JSON format:
{
  "subject": "...",
  "aspect": "...",
  "verb_object": "...",
  "explanation": "..."
}
--- Example ---
Input Text:
i have started using deepl for translations and it's so much better than google
translate it's made its way up into my top most used websites
Output JSON:
{
  "subject": "DeepL",
  "aspect": "translation quality compared to Google Translate",
  "verb_object": "outperform Google Translate",
  "explanation": "The speaker reports that DeepL provides higher translation quality
  than Google Translate and has become one of their most frequently used tools."
}
--- End Example ---
\end{Verbatim}
\end{framed}

\section{Detailed Evaluation Breakdown}
\label{sec:appendix_eval_detail}

Table~\ref{tab:evaluation_detail} reports per-community community-label accuracies for the stratified validation subset (100 items per community). Labels were independently checked by two annotators. When annotators disagreed, we applied a fixed adjudication rule rather than discussion based resolution: the first annotator's label was retained as the final reference label.

\begin{table}[H]
\centering
\small
\begin{tabular}{lr}
\hline
\textbf{Community} & \textbf{Accuracy} \\
\hline
AI developers & 0.72\\
Language learners & 0.77\\
Language service providers & 0.74\\
Professional translators & 0.93\\
Unclear & 0.97 \\
\hline
\end{tabular}
\caption{Per-community accuracy for community-label validation on a stratified sample of 500 items, with 100 items per community.}
\label{tab:evaluation_detail}
\end{table}
To examine whether community classification performance varies across platforms, we additionally evaluate the classifier on a platform-stratified validation subset.
Table~\ref{tab:platform-community-accuracy} reports accuracy by platform.
The results show that classification accuracy is highest on Reddit and Bluesky, lower on Mastodon, and lowest on Facebook.
This variation likely reflects differences in discourse style, metadata structure, and the degree to which community identity is explicitly signaled across platforms.
\begin{table}[H]
\centering
\small
\begin{tabular}{l c}
\toprule
\textbf{Platform} & \textbf{Accuracy} \\
\midrule
Reddit   & 92.16\% \\
Bluesky  & 91.30\% \\
Mastodon & 80.00\% \\
Facebook & 72.83\% \\
\bottomrule
\end{tabular}
\caption{Platform-specific accuracy for the community classifier.}
\label{tab:platform-community-accuracy}
\end{table}

\section{Platform Composition by Year}
\label{sec:appendix_platform_time}

Table~\ref{tab:platform_composition_year} reports counts and percentages by platform and year. This supports interpretation of Figure~\ref{fig:fig1}: growth in post volume over time may partly reflect increased representation of Facebook in later years rather than a uniform increase across platforms.

\begin{table}[H]
\centering
\scriptsize
\caption{Platform composition by year (percentages with counts in parentheses).}
\label{tab:platform_composition_year}
\resizebox{\linewidth}{!}{
\begin{tabular}{lrrrr}
\hline
\textbf{Year} & \textbf{Facebook} & \textbf{Reddit} & \textbf{Bluesky} & \textbf{Mastodon} \\
\hline
2019 & 93.2\% (5,280) & 6.4\% (361) & 0.0\% (1) & 0.5\% (26) \\
2020 & 89.6\% (6,830) & 10.0\% (765) & 0.0\% (2) & 0.4\% (29) \\
2021 & 90.5\% (6,808) & 8.9\% (667) & 0.0\% (3) & 0.6\% (43) \\
2022 & 89.7\% (8,657) & 7.2\% (698) & 0.0\% (3) & 3.0\% (290) \\
2023 & 72.3\% (9,593) & 22.6\% (2,996) & 1.1\% (143) & 4.0\% (531) \\
2024 & 75.9\% (14,933) & 18.8\% (3,699) & 3.5\% (688) & 1.9\% (366) \\
2025 & 64.5\% (24,330) & 21.8\% (8,210) & 12.3\% (4,627) & 1.4\% (540) \\
\hline
\end{tabular}
}
\end{table}

\section{Evaluation Details for Automatic Community Labels}
\label{sec:appendix_community_eval}
\begin{table}[H]
\centering
\scriptsize
\resizebox{\linewidth}{!}{
\begin{tabular}{lrrrr}
\hline
\textbf{Class} & \textbf{Precision} & \textbf{Recall} & \textbf{F$_1$} & \textbf{Support} \\\hline
AI developers & 0.889 & 1.000 & 0.941 & 72 \\
Language learners & 0.986 & 1.000 & 0.993 & 68 \\
Language service providers & 1.000 & 1.000 & 1.000 & 72 \\
Professional translators & 1.000 & 1.000 & 1.000 & 93 \\
Unclear & 1.000 & 0.908 & 0.952 & 109 \\\hline
\end{tabular}
}
\caption{Per-class precision, recall, and F$_1$ for community labels on the manually validated subset.}
\label{tab:community_prf}
\end{table}
Table~\ref{tab:community_prf} summarises per-class precision, recall, and F$_1$ for the same evaluation set.

\section{Conflict Metrics and Trend Definitions}
\label{sec:appendix_conflict_metrics}

This appendix provides the formal definitions of the within-community conflict metric and the expansion/escalation labels used throughout our analysis (Figures~\ref{fig:fig2}).

\paragraph{Within-Community Conflict}
For a given (community, year, aspect) cell with $n$ posts and real-valued sentiment scores $s_1, \ldots, s_n$, we compute the sample mean and standard deviation
\[
\begin{aligned}
  \bar{s} &= \frac{1}{n} \sum_{i=1}^{n} s_i, \\
  \sigma &= \sqrt{\frac{1}{n-1} \sum_{i=1}^{n} \left(s_i-\bar{s}\right)^2}.
\end{aligned}
\]
Let $p_{\text{pos}}$ and $p_{\text{neg}}$ be the proportions of posts with positive and negative sentiment scores, respectively:
\[
\begin{aligned}
  p_{\text{pos}} &= \frac{1}{n} \sum_{i=1}^{n} I(s_i > 0), \\
  p_{\text{neg}} &= \frac{1}{n} \sum_{i=1}^{n} I(s_i < 0).
\end{aligned}
\]
We define a polarity mixing term
\[
  \text{polarity\_mix} = 4\, p_{\text{pos}} \, p_{\text{neg}} \in [0, 1],
\]
and the within-community conflict metric as
\[
  \text{conflict\_within} = \sigma \cdot \text{polarity\_mix}.
\]
This metric is high when sentiment polarity is both \emph{spread out} (large $\sigma$) and \emph{mixed} (substantial mass on both positive and negative labels). It captures polarity-based dispersion within a community--topic--year cell, not argumentative conflict or contested norms per se; we use it as a heuristic to flag topics where evaluative stance is most divided at the level of sentiment.
\section{Within-Community Dynamics: Conflict and Change}
\label{sec:appendix_within_community}

This appendix provides extended supporting detail for the within-community findings summarized in \S\ref{sec:analysis}. We focus on two forms of within-community variation: (i) \emph{within-community conflict} (operationalised as sentiment-polarity dispersion; see \S\ref{sec:appendix_conflict_metrics}), and (ii) \emph{within-community change over time}, where discussion priorities shift as tools and practices evolve.

\subsection{Within-Community Conflict Patterns}

We quantify within-community conflict as the degree to which members of the same stakeholder group express divergent attitudes toward the same technology--topic combination. For each combination, we compute a \emph{within-community conflict} score that combines sentiment dispersion with the co-occurrence of strongly positive and strongly negative evaluations (formal definition in \S\ref{sec:appendix_conflict_metrics}).

To identify high-conflict topics, we examine conflict scores for each community over time. Figure~\ref{fig:fig2} shows that high-conflict cells cluster in a relatively small set of aspects---notably translation quality, features and functionality, and efficiency/cost. These hotspots are most pronounced among professional translators and language service providers, but elevated conflict also appears among language learners (especially around social and economic impacts) and AI developers (especially around reliability and trust). We use these metrics to identify candidate ``conflict zones'' for qualitative analysis.

\paragraph{Qualitative illustration: AI developers on Reliability \& Trust}

To make our sentiment-based definition of within-community conflict concrete, we examine AI developers' discussions of \emph{Reliability \& Trust}.
Here, conflict refers to the coexistence of positive and negative evaluations within the same community and topic, rather than to disagreement between different stakeholder groups.
Positive posts frame MT and LLM-based translation as robust, useful, or reliable in bounded settings, while negative posts emphasize hallucinations, omissions, verification needs, API instability, and limits of probabilistic generation.
Table~\ref{tab:ai_dev_reliability_mixed} provides representative snippets from both sides.
These examples show why \emph{Reliability \& Trust} is internally contested among AI developers: the same technical community contains both confidence in engineering solutions and skepticism about whether MT systems are trustworthy enough for real translation workflows.

\subsection{Within-Community Change Over Time}

Not all within-community variation is conflict; communities also exhibit \emph{shifts in focus} as technologies and practices change. To trace these shifts, we examine professional translators' discussions of Features \& Functionality from 2021 to 2025. Table~\ref{tab:features_functionality_timeline} summarises the changing yearly focus and associated tensions.

Across these stages, translators consistently evaluate new features through the lens of workflow efficiency and professional control. They welcome tools that reduce friction, but raise concerns when automation or ``feature-rich'' updates reduce their ability to supervise outputs or disrupt established workflows. This pattern illustrates within-community \emph{change} in focus---from CAT infrastructure to AI integration and workflow automation---as distinct from within-community \emph{conflict}, which captures simultaneous divergence in evaluative stance.

Taken together, the conflict case (AI developers on reliability and trust) and the change-over-time case (professional translators on features and functionality) show how the same broader technical transition---the mainstreaming of LLM-based translation---produces different forms of variation across communities: sharp internal disagreement in some contexts and evolving discussion priorities in others.

\begingroup
\small
\setlength{\tabcolsep}{4pt}
\renewcommand{\arraystretch}{1.12}

\begin{table*}[t]
\centering
\footnotesize
\begin{tabular}
{@{}P{0.15\textwidth}P{0.39\textwidth}P{0.39\textwidth}@{}}

\toprule
\textbf{Language pair} &
\textbf{Positive views} &
\textbf{Negative views} \\
\midrule

Japanese $\rightarrow$ English
& \textbf{AI developers:} \emph{``It's definitely better ... when translating Japanese to English, it's more accurate.''}\par
  \textbf{Professional translators:} \emph{``Neither is good enough without careful checking and rewriting, but they do enable me to work much faster.''}
& \textbf{AI developers:} \emph{``Japanese to English machine translation can be hilarious ... the computer will then wildly guess.''}\par
  \textbf{Professional translators:} \emph{``AI models like DeepL are still almost completely useless. With complex literary language, they just fail.''} \\

\midrule

English $\rightarrow$ Spanish
& \textbf{Language learners:} \emph{``I use a Chrome extension DeepL for English to Spanish translation and it seems to do a good job.''}\par
  \textbf{Language service providers:} \emph{``English to Spanish is very good ... I ask it to adapt idioms and expressions so they sound natural.''}
& \textbf{Language learners:} \emph{``I'm trying to translate a PDF from English to Spanish ... but ChatGPT always returns a single-page summary.''}\par
  \textbf{AI developers:} \emph{``The text translated by the NLLB model tends to miss details, and the accuracy of the slang isn't great.''} \\

\midrule

English $\rightarrow$ Arabic
& \textbf{AI developers:} \emph{``When translating from English to Arabic, Claude 3 Opus provides the best results, resembling an expert human translator.''}\par
  \textbf{AI developers:} \emph{``Gemini [is] more effective than ChatGPT and Claude for translating AI messages from English to Arabic.''}
& \textbf{Language learners:} \emph{``DeepL ... doesn't support Arabic as far as I know. My main languages are English--Arabic.''}\par
  \textbf{Language service providers:} \emph{``Google [gives] the wrong translation, which is misleading ... the medical term was rendered literally.''} \\

\midrule

Arabic $\rightarrow$ English
& \textbf{Language learners:} \emph{``If you've ever tried using [Google Translate] to translate Arabic into English, the results can be amazing.''}\par
  \textbf{AI developers:} \emph{``Any recommended LLMs that would do a good job in translating ... Arabic to English on an M1 Max Mac?''}
& \textbf{Language learners:} \emph{``Arabic to English ... I don't think Google Translate is being all that accurate.''}\par
  \textbf{Professional translators:} \emph{``Do not use Google to translate Arabic to English when complex meanings are involved.''} \\

\midrule

English $\rightarrow$ French
& \textbf{AI developers:} \emph{``The language combinations English--German and English--French are very good, definitely better than Google Translate.''}\par
  \textbf{Language service providers:} \emph{``Google has started rolling out live translated captions for Google Meet ... from English to French, German, Portuguese and Spanish.''}
& \textbf{Language learners:} \emph{``They do the same with French, translating English to French, but the translation is ... garbage.''}\par
  \textbf{AI developers:} \emph{``I can't believe they can't get LLM translation to work English to French. Those are ... two very high-resource languages.''} \\

\bottomrule
\end{tabular}
\end{table*}
\endgroup

\clearpage

\begingroup
\small
\setlength{\tabcolsep}{4pt}
\renewcommand{\arraystretch}{1.12}

\begin{table*}[t]
\centering
\footnotesize
\begin{tabular}{@{}P{0.15\textwidth}P{0.39\textwidth}P{0.39\textwidth}@{}}

\toprule
\textbf{Language pair} &
\textbf{Positive views} &
\textbf{Negative views} \\
\midrule

Spanish $\rightarrow$ English
& \textbf{AI developers:} \emph{``As a bilingual Spanish/English speaker, I think ChatGPT is phenomenal at translation. I can even ask for localized dialects.''}\par
  \textbf{Professional translators:} \emph{``I translated the lyrics to a Spanish rap song ... and GPT gave me a fine translation. It only needed a couple edits.''}
& \textbf{Language learners:} \emph{``Spanish to English ... Google Translate is failing me.''}\par
  \textbf{Language learners:} \emph{``Does anyone have problems in DeepL online where it just omits full paragraphs? I'm translating from Spanish to English.''} \\

\midrule

Chinese $\rightarrow$ English
& \textbf{AI developers:} \emph{``Chinese to English. Not just better than Google, but the best, by far, of everything I have tried.''}\par
  \textbf{Professional translators:} \emph{``I'm bilingual and have been translating Chinese to English texts; it does a bang-on job, much better than Google or DeepL.''}
& \textbf{AI developers:} \emph{``Google Translate would sometimes drop words when going from Chinese to English.''}\par
  \textbf{AI developers:} \emph{``From my usage of Chinese to English translation, Gemma 3 27b was one of the worst models I've used for the task.''} \\

\midrule

German $\rightarrow$ English
& \textbf{Language service providers:} \emph{``I have been very satisfied with the sophistication that DeepL provides for German--English.''}\par
  \textbf{Professional translators:} \emph{``I have tried translating poems from German to English and vice versa. The AI has done a great job.''}
& \textbf{Language learners:} \emph{``I use it for German--English. In Google Translate grammar and capitalization is all wrong.''}\par
  \textbf{AI developers:} \emph{``For German to English it works wonderfully, especially in legalese ... For casual or creative texts, though, it's no good.''} \\

\bottomrule
\end{tabular}
\caption{Representative post/comment snippets for selected language-pair contexts, grouped by community label. The excerpts illustrate mixed sentiment and use cases in language-pair discussions; they should not be read as a comprehensive evaluation of MT performance for any language pair. Snippets are shortened and lightly cleaned by removing URLs, handles, and non-essential context.}
\label{tab:language-pair-case-studies}
\end{table*}
\endgroup

\begin{table*}[t]
\centering
\footnotesize
\setlength{\tabcolsep}{3.5pt}
\begin{tabular}{@{}p{0.15\textwidth}p{0.25\textwidth}p{0.34\textwidth}p{0.20\textwidth}@{}}
\toprule
\textbf{Topic} &
\textbf{AI Community (subtopics)} &
\textbf{Non-AI Community (subtopics)} &
\textbf{Framing Gap} \\
\midrule

\textbf{Translation Quality} &
Accuracy and adequacy; context, nuance, and register; human review and quality assurance &
Human review and quality assurance; accuracy and adequacy; context, nuance, and register &
Measurable quality vs. accountable quality \\

\textbf{Efficiency} &
Cost, tokens, and compute resources; workflow integration; model optimization and inference efficiency &
Workflow integration; MTPE / post-editing workflow; quality control and review burden &
System efficiency vs. human workload \\

\textbf{Reliability \& Trust} &
Accuracy, hallucination, and factual trust; trust, adoption confidence, and reputation; stability, outages, and bugs &
Trust, adoption confidence, and reputation; accuracy, hallucination, and factual trust; verification, review, and human oversight &
Technical reliability vs. practical trust \\

\textbf{Cost \& Pricing} &
Free, discounted, and low-cost access; subscription, licensing, and usage plans; token, compute, and API cost &
Per-word, per-project, and freelance rates; free or low-cost access; market pressure and price competition &
Infrastructure cost vs. labour pricing \\

\textbf{Integration \& Workflow} &
APIs, plugins, and automation hooks; file exchange and format handling; project management and collaboration workflow &
CAT tools, TMS, and platform integration; file exchange and format handling; project management workflow &
Technical integration vs. production workflow \\

\textbf{User Adoption \& Social/Economic Impact} &
Income, pricing, and market pressure; hiring and freelance work; adoption, trust, and acceptance &
Income, pricing, and market pressure; hiring and freelance work; productivity, business value, and competitiveness &
Market adoption vs. labour consequences \\

\bottomrule
\end{tabular}
\caption{Gaps between the AI community (AI developers) and non-AI communities (language learners, LSPs, and translators).
For topic, we list subtopics from both AI and non-AI communities, based on our corpus (2023--2025).
}
\label{tab:subtopic_framing_gap_full}
\end{table*}

\end{document}